\documentclass{article}
\usepackage{ijcai25}

\usepackage{times}
\usepackage{soul}
\usepackage{url}
\usepackage[hidelinks]{hyperref}
\usepackage[utf8]{inputenc}
\usepackage[small]{caption}
\usepackage{graphicx}
\usepackage{amsmath}
\usepackage{amsthm}
\usepackage{booktabs}
\usepackage{algorithm}
\usepackage{algorithmic}
\usepackage[switch]{lineno}

\title{Graph-Based Orchestration of Heterogeneous AI Models:\\ A Control Node Approach to Composite Intelligence}

\author{
G. Michael Youngblood
\and
Filip Dvo{\v{r}}{\'a}k\and
Slavomir Svancar\and
\\Tom{\'a}{\v{s}} Balyo\and
Michal Ficek\And
Martin Dousek\\
\affiliations
Filuta AI, Inc., 1606 Headway Cir STE 9145, Austin, TX 78754, USA\\
\emails
\{michael, filip, slavo, tomas, michael, martin\}@filuta.ai
}

\begin{document}

\maketitle

\begin{abstract}
We present a graph-based framework for Composite AI systems that integrate multiple AI techniques to address complex problems. We describe an architecture using specialized control nodes to orchestrate diverse models including planning, machine learning, constraint programming, and large language models. The system employs directed acyclic graphs and behavior trees to manage model relationships and execution sequencing. Our methodology encompasses data connectors, computational services, and a containerized deployment framework supporting hierarchical, sequential, and concurrent model composition patterns. Through case studies, we demonstrate how this architecture enables the creation of composite systems that leverage the strengths of multiple AI techniques within a unified operational schema, overcoming the limitations of single-model approaches.
\end{abstract}

\section{Introduction}

There are many techniques for modeling real-world problems, and no single technique will dominate as the most efficient across most real-world problems for the near future. Instead, modeling real-world problems can be seen as a composition of chained, concurrent (parallel or interleaved), and/or nested techniques/models, using the most efficient technique/model for each of the sub-problems into which the real-world problem decomposes. Significant validation of this composition approach to solving industrial problems can be found in the studies performed by Gartner \cite{Gartner2022}, where Composite AI has emerged as one of the most promising techniques, following the under-performance of deep learning and machine learning in applied projects.

The system and methods presented here provide a common ground for the composition of various models and techniques, which we will just refer to simply as just “models” going forward, into a system. These models include AI Planning, machine learning, scheduling, graph theory, constraint programming, linear programming, large language models (LLMs) such as GPT \cite{Floridi2020}, multimodal foundational models, and any other type of element that using computation can perform a distinct set of functions producing output from input.

This work is motivated by creating systems composed of many models to address the challenges single models face in being able to perform many disparate functions outside of their design/training. Models cannot reason about the optimal plan of actions to achieve a goal, dispatch and monitor action execution, act on state information, diagnose failures, and adapt to unexpected events all in a single model. Current automation challenges also involve connecting all actors in the system and collecting their real-time operational data. Efficient and reliable control of a target system to achieve goals requires reasoning about this data using an appropriate composition of models leading to appropriate action decisions.

We introduce a methodology including key building blocks for constructing automation systems designed as a graph with the flexibility to transform into many graph forms for deployment. The key architectural invention is the introduction of a data and model aware, logic-driven control node central to defining a composition that unifies a plurality of disparate models. These compositions can be used to generate code, be containerized, and then deployed autonomously.

The Composite AI system provides capabilities to incorporate individual models, allowing the human user to augment the models, compose the models into the operational schema with well-defined inputs, outputs, and goals, and generate stand-alone intelligent automation services that act towards user-defined goals based on the model composition and observations of the world.

\section{Graphs}
Our Composite AI system's foundation is a graph structure, which models pairwise relationships between entities through vertices (nodes) and edges (links). This representation offers significant implementation advantages, including clear visualization of relationships, efficient algorithmic processing, modular design, and adaptability to dynamic changes. Graphs effectively capture complex hierarchical structures, aid in problem decomposition, and support various data storage formats, making systems more efficient and scalable.

\subsection{Directed Acyclic Graph (DAG)}
A Directed Acyclic Graph (DAG) features directed edges with specific orientation and contains no cycles. This structure excels at managing dependencies and order, ensuring tasks execute only after prerequisites are met. The absence of cycles eliminates circular dependencies, simplifying resolution and preventing deadlocks. DAGs support efficient optimization algorithms and provide clarity in representing complex systems, making them ideal for system management and control.

\subsection{Behavior Trees (BT)}
Behavior trees are hierarchical graphs that define autonomous agent behavior in a tree-like structure. They consist of three node types:
\begin{itemize}
    \item Composite nodes (→ and ?) control execution flow by organizing other nodes
    \item Decorator nodes modify child node behavior
    \item Leaf nodes perform actions or checks
\end{itemize}

\begin{figure}[h]
    \centering
    \includegraphics[width=0.5\textwidth]{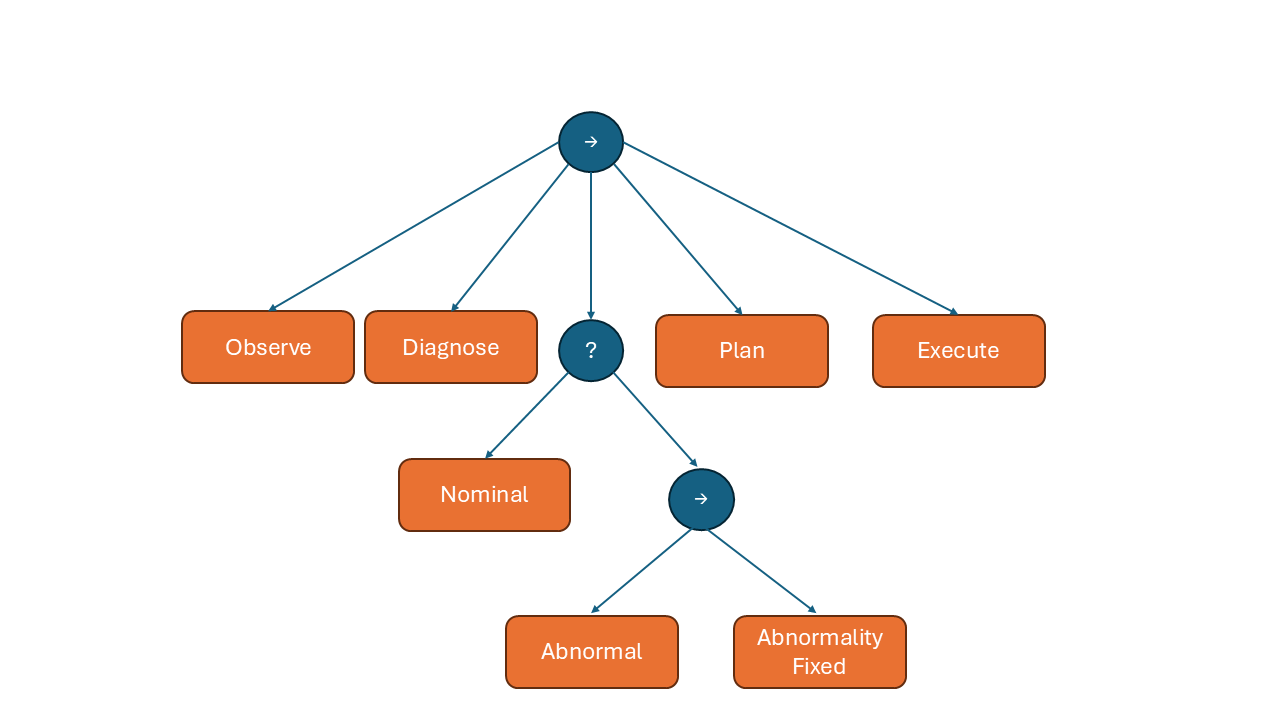}
    \caption{Example Behavior Tree}
    \label{fig:behavior_tree}
\end{figure}

Behavior trees enable hierarchical problem decomposition, allowing higher-level trees to incorporate lower-level trees without requiring detailed knowledge of subproblems. They can integrate planning and diagnosis capabilities, with each action represented as a node whose evaluation reflects the action's execution status. Within the framework, behavior trees function as deterministic policies that observe the world and initiate appropriate actions.  Figure~\ref{fig:behavior_tree} illustrates an example behavior for monitoring and adapting a plan in execution. In the example, blue nodes have the following behavior:
\begin{itemize}
\item → (and) evaluates to: 
\begin{itemize}
\item success if all its children evaluate to success,
\item failure if at least one node evaluates to failure, and
\item running if at least one node evaluates to running while none evaluates to failure.
\end{itemize}
\end{itemize}
\begin{itemize}
\item ? (or) evaluates to:
\begin{itemize}
\item success if at least one child evaluates to success,
\item failures if all children evaluate to failures, and
\item running if at least one child evaluates to running and none evaluates to success.
\end{itemize}
\end{itemize}
\vspace{-0.2in}

The key advantages of behavior trees include verifiability, human-readability, and visualization capabilities, providing a common platform for sharing architectural concepts and domain knowledge among stakeholders.



Within a framework, a behavior tree can be seen as a (deterministic) policy that observes the world with each tick of the tree and may send actions, such as “Plan.” Planning can be a subproblem itself, implemented as a policy through a solver that takes state, actions, and a goal to develop a plan for execution.

Behavior trees are verifiable, human-readable, and easy to visualize. They provide a common ground for sharing ideas, architectures, and domain knowledge among domain experts, product managers, and AI systems.

\section{Methodology}

Learning a single end-to-end model is rarely computationally feasible. Instead, end-to-end models are typically composed of multiple heterogeneous models, some of which are machine-learned while others are human-built using AI and operations research modeling techniques. The system and methods described in this paper support the design, development, and deployment of these composite AI systems.

\begin{figure*}[t]
    \centering
    \captionsetup{justification=centering}
    \includegraphics[width=0.7\textwidth]{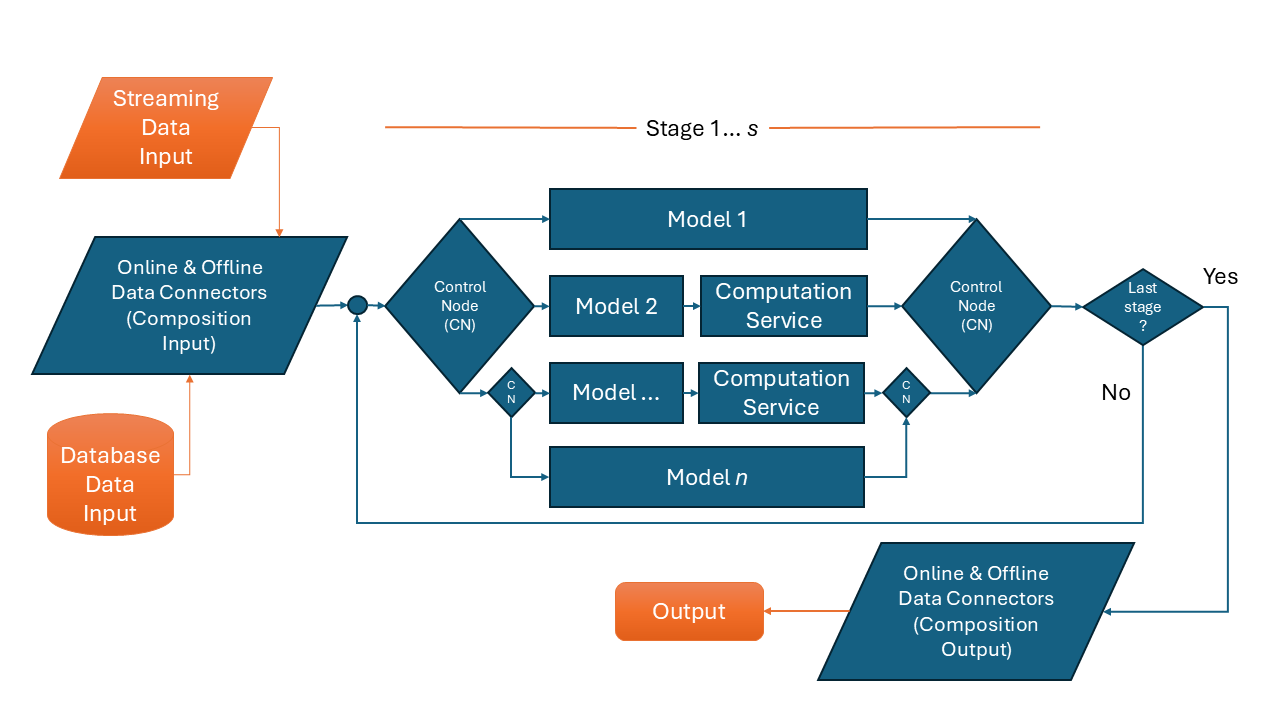}
    \caption{Composition framework. Each control node is unique being configured for their specific\\ downstream components through introspection or manually, but use a common code base.}
    \label{fig:comp_framework}
\end{figure*}

The system goal is the automated assembly of data connectors, domain and goal-based problem models, computational services (e.g., solvers, reasoners, large-language models, etc.), and appropriately configured control nodes (logic-gated demultiplexers, multiplexers, and pass-through using data and model knowledge) using a graph-based organization into a service supported by an externally interfacing API that can be plugged into a target system for automation as a Composite AI system. The Composition Framework shown in Figure~\ref{fig:comp_framework} illustrates an abstract graph of the principal components in making compositions (the blue parts). Online and/or offline data is brought into the system and adapted for distribution to one or more models organized into one or more stages. Novel control nodes can flexibly be configured to distribute data into and out of model in to, between, and out of stages. As shown, some models require computational pipelining. At the end of all stages the composition produces output. This composite core is wrapped in a service API and designed for deployment in a container or set of containers to provide the configured service both in a testing and production environment.

\subsection{Online and Offline Data Connectors }
When building real-time intelligent services, it is necessary to process both offline and online data. Offline data, which are typically used for model learning, can be quite large and are not significantly constrained by processing time. In contrast, online data are expected to arrive in real-time when the service is deployed, and the results of processing the real-time data are going to be produced and published by the service.

While the connectors are critical components of the platform and composed services, the implementation effort has already been invested in by several open-source data connectors, such as Airbyte \cite{Airbyte2023}. The system builds upon an existing framework of connectors developed by others and extends their functionality by adding preprocessing data transformations relevant to learning and/or deploying the models.

\subsection{Models}
The core workers in a composite AI system are the models. Models may be pre-trained with data, augmented with new data, learn from zero using new data, or only algorithmic in nature (i.e., a technique). When models need to be augmented or developed from the ground up then the model-specific techniques apply such as Automated Domain-driven Model Synthesis \cite{balyo2024planning}. By applying various filters and transformations to a dataset ingested from a data connector, the system (optionally aided by a user) can define the state space or other features, which are then used to learn the model. The same dataset can be used to define state spaces or inputs for multiple different modeling approaches. For example, we can model a predictive task as a dependency of the engine temperature on speed, road incline, and total weight, while simultaneously modeling vehicle logistics using classical planning \cite{Ghallab2004}.

\subsection{Computational Services}
Some models require computational support to execute. For example, a planning model needs to be run through a solver with a problem specification. A GPT (Generative Pre-trained Transformer) model needs to have its weights loaded into the configured transformer architecture and be run with a problem prompt. One case study of the system supports modeling with classical planning using machine learning \cite{Dvorak2021}, as well as constraint modeling using CP-SAT in \cite{Google2022} and machine learning \cite{LeDell2020}.

\subsection{Control Nodes: Logic-Gated Multiplexers, Demultiplexers, and Pass-throughs}
The key to creating complex composite AI systems is providing a control mechanism to combine, select, transform, deconstruct, and configure how the data flows into and out of the models in each stage until final output. Each control node is configured to marshal the data into or out of a set of models and configure it as required using the following configuration elements.

\begin{enumerate}
    \item \textbf{Permeability}: Defines how data flows through the gate.
    \begin{enumerate}
        \item Fully permeable: Data is a complete pass-through to one or more models downstream.
        \item Partially permeable: Data is a partial pass-through to one or more models downstream and may be combined, selected, or transformed before being passed to the model(s).
        \item Impermeable: Data is not passed-through and will be combined, selected, or transformed before being passed to one or more models downstream if at all.
    \end{enumerate}
    \item \textbf{Combination}: Define how inputs are combined before being passed to one or more models downstream. 
    \item \textbf{Selection}: Define how selected inputs are passed to one or more models downstream if at all (e.g., filtered). 
    \item \textbf{Transformation}: Define how inputs are altered before being passed to one or more models downstream.
    \item \textbf{Downstream Model Configuration}: Define specific model configurations for each model as needed for each input. For example, each input into a LLM may specify some specific hyper-parameters associated for processing the input.
    \item \textbf{Orchestration}: Defines conditions of received outputs and when/how to proceed in processing. Orchestration may only require a few responses instead of all responses from a set of upstream models before proceeding. It may require specific properties to the responses before proceeding. However, it should never permanently halt processing and any resolutions needed to proceed should be able to be kicked off by the gate.
\end{enumerate}

\begin{figure}[h]
    \centering
    \includegraphics[width=0.5\textwidth]{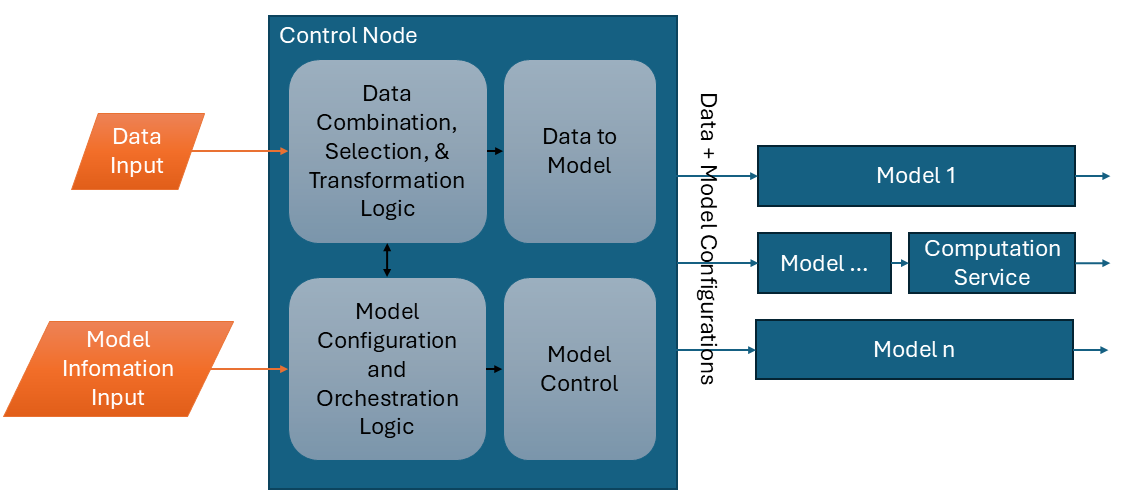}
    \caption{Control Node Architecture}
    \label{fig:control_node}
\end{figure}

In a typical case study as shown in Figure~\ref{fig:control_node}, the control node is implemented in a computer programming language into four areas of functionality. Data comes into a Data Combination, Selection, and Transformation Logic component that applies logic rules to the incoming data to combine, select (filter in or out), and transform the data as specified. A Model Configuration and Orchestration Logic component receives information about the models it can use and is configured through logic rules on how to set up, execute, and orchestrate the downstream models. These two components may share information that informs logic rules for processing. The Data to Model component provides the data to the appropriate downstream component(s), which may be the system output. The Model Control component configures the downstream models, if they exist, and kicks off their operation as defined by the logic rules. Control Nodes are aware through introspective or manual configuration of their downstream components and parameters. Control nodes always have incoming data and should have some outgoing data but may not have any model input or control. Any logical specification system can be used to implement a control node.

One example of how control nodes are used in a stage is to reduce LLM hallucinations, generated incorrect or fabricated information, by passing the same prompt into each of a set of LLM models (e.g., ChatGPT, Llama, and Gemini) in parallel. The outputs are then transformed into an intersecting set so that only outputs that are in fuzzy match are passed to the next stage. Information that is not the same in each response is filtered reducing inconsistent aspects, which are often manifested as LLM hallucinations.

Control nodes can output to other control nodes and be nested in model flows as shown in Figure~\ref{fig:behavior_tree}. Control nodes fit neatly into the system graph as nodes before and after models and model processing nodes. Most times system control is best defined as a DAG or BT, but in some case studies a directed graph may be used with cycles. These cycles typically specify a loop between control nodes so that information may be processed multiple times by the same set of models. 



Compositions may be designed to operate under various modalities. One common set is the train and test (i.e., learn and perform, explore and exploit). Control nodes can be specified with different logic configurations to support each desired mode of composition usage (e.g. using environmental variables). 

\section{Architectural Solution Composition}
Architectural Solution Composition in practice involves building a graph (behavior tree in this example) that combines multiple models with corresponding solvers, goals, and APIs into a service.

Although the models share a single observation space, each model only considers its own local state space. Each model can be tied to a computation service (e.g., solver), which may continuously work on problem instances generated from the observations whenever it is triggered. The system can use any type of software defined computation services including planning solvers in the Unified Planning Framework \cite{unified_planning_softwarex2025} and the wide range of AI techniques.

We often encapsulate reasoning and action execution into the nodes of a behavior tree providing the grounds to define a composed automated reasoning, planning, and execution system in terms of tree composition. The tree composition is an operationalization of multiple models together with the corresponding solvers, actors, goals, and APIs into a service. Any two or more models in a composition are related either by 

\begin{enumerate}
    \item nesting (hierarchical), when one model is fully nested within another, and the subsystem operating with the outer model can invoke the subsystem operating the inner model, 
    \item precedence or chaining (sequential), when a subsystem for the earlier model can be fully executed before running the subsystem for the later model, or 
    \item concurrent (interleaving and/or parallel), when subsystems for both models can run concurrently and interfere with each other, for example by exchanging their best solutions.
\end{enumerate}

These three model composition approaches are available to the solution architect through the control nodes, and it is upon the architect to design a composition appropriate for the given real-world problem. The models share a single observation space, the control nodes picking only the data relevant for them (e.g., in a food delivery system example, vehicle routing with time windows precedes running a planner). We first produce optimal assignments of orders to drivers, which is equivalent to adding control knowledge to the more general planning problem, which is solved consequently. We consider each model to be tied with a computational service subsystem as appropriate (e.g., plan with a planner, problem solve with a solver), which keeps solving the problem instances generated from the observations every time it is triggered.

The concept of model composition in the system is another tool typically provided to the solution architect user. If desired, the user can build and compose models to guarantee optimal solutions. There can also be cases when optimality is infeasible or not desired, and the goal is to find solutions quickly with a reasonable distance from optimality.

Service code can be generated by introspectively filling out template code of various components of the composition, or by configuring pre-build services (e.g., using JSON configuration files, environment variables, or API calls). In implementation there typically exists a code/service template for every type of composition node. The solution is composed of a main orchestration service, that contains knowledge about all the components in the composition. The individual nodes are either directly part of the orchestration service, in case of simple nodes, or the orchestration service has lightweight proxy wrappers for a dedicated microservice that are logically part of the composition. The orchestration service communicates with the microservice using this proxy and can either directly fetch and push data to it via this proxy or can use it to configure the wrapped service to specify data inputs and outputs. Individual nodes are decoupled, language-independent, and are flexible to be able to be part of a wide range of compositions. 

All the parts of the solution templates, including the orchestration service and all the engineered packages should be packed into container technology (i.e., Docker) by design. After code generation, all that is needed is to execute the already created build commands to create the containers and service manifest. These containers then can be easily versioned, duplicated, and deployed either locally or to a cloud solution, and can be used in various environments, such as testing, staging, demo, or production. 

The Filuta AI system implements all of these aspects into a commercial platform used to provide solutions to real-world problems such as video game quality assurance evaluations. 

\section{Deployment}
Deploying services produced by the system is a push-button action for the user. It involves a cloud-agnostic or computational platform deployment of containerized microservices at the backend. Once deployed, users can monitor the deployment status and interact with it at the defined network address through the API defined in the composition.

Final deployment also contains implicit and explicit tools and services not directly part of the data pipeline. These services allow us to incorporate functional and non-functional requirements to the solution, such as High Availability, extensive monitoring and logging capacities, authorization, and authentication mechanisms. Tools such as monitoring are implicitly part of every deployment and are not visualized in the Composition UI representation. An example of explicit supporting tools is the High Availability Service that will deploy the Composition multiple times in multiple regions to serve as a failback if one instance fails, to allow faster and more scalable usage for customers from different regions, or for many customers in one place.

Deployed services can be used in two separate ways:
\begin{itemize}
    \item Dedicated for a composition to which they belong
    \item Distributed across multiple service environments, customer solutions, or even multiple customers.
\end{itemize}

Dedicated services are, for example, models dependent on the environment where they are used or if the customer wants dedicated services running for them. Distributed services can be either shared data sources that are the same for all the customers, for example a weather service, or resource-heavy services that would be cost prohibitive to use for a single deployment, such as an LLM service. Customer isolation is always guaranteed by the solution either by using dedicated services or by not sharing context in the distributed services.

\begin{figure*}[t]
    \centering
    \includegraphics[width=0.52\textwidth]{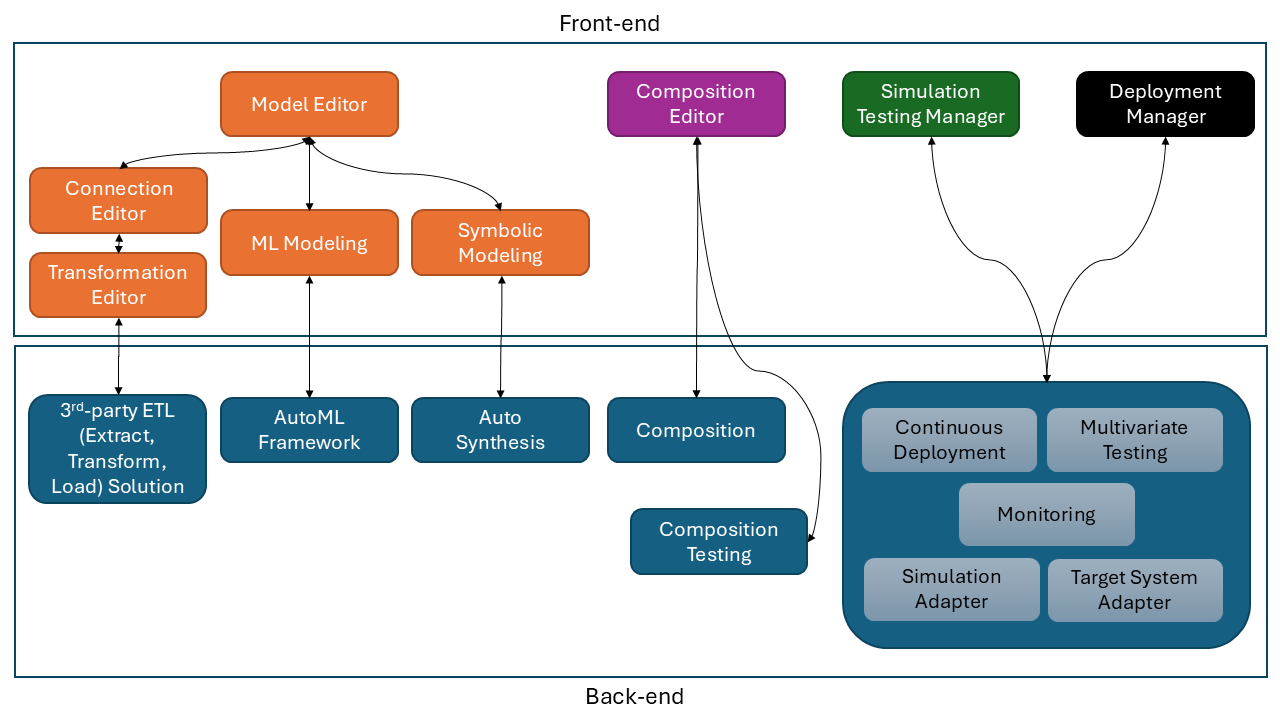}
    \caption{An end-to-end system for Specification, Codification, and Deployment of Composite AI Systems}
    \label{fig:end-to-end}
\end{figure*}

Composition is the core part of a complete system as shown in Figure~\ref{fig:end-to-end} with four main components:

\noindent 1. \textbf{Model Editor} allows the expert user to define data inputs, and how they are transformed into an ingestible format. For example, tabular data for predictors/ML models, or time series for symbolic models. It then provides a way to plug those connectors/transformers into an ML or symbolic modeling approach. It is possible for the expert user to interactively adjust and augment the proposed model until they are satisfied with its quality. This could mean accuracy for ML, or human-check and test validation against input data for symbolic models.

Models are used in Compositions and have the following properties: 1) Defined input specifications, 2) Defined output specifications, 3) Defined model configuration parameters, 4) Defined model controls, 5) Defined computation and memory requirements, and 6) Characterization of performance on data.

\noindent 2. \textbf{Composition Editor} takes multiple models and offers diverse options to combine them using control nodes to implement a graph that can be executed as a DAG, behavior tree, or other forms as well as using data connectors to provide shared observation space. The editor encodes both the decomposition of the mathematical structure representing the real-world problem and the operationalization of the model with the behavior of a general agent. This feature allows for quick iteration, learning, and versioning of agents that provide intelligent automation services. The editor also specifies I/O.

The building blocks of a composition are explicitly versioned to support breaking changes—if one service changes their API and other dependent service needs to adjust, the system can keep one composition that contains services that have not adjusted yet on an old version and another composition that already has all the changes incorporated at a newer version. This also supports falling back to older versions if needed, rolling out features gradually, and other cases where explicit versioning supports better system management. 

The building blocks are used in case studyies to 1) construct the behavior tree or other executable graph in Python (or another OOP language based on customer needs, such as C++ for game testing integrations) and 2) list the deployment requirements of the components necessary to run the composition. The behavior tree represented by the composition in a case study has shown to be a low-latency (1-5ms), minimal-computation tree structure that constantly runs when deployed and manages the other components (containers) considered to be serviced in this interaction.

\noindent 3. \textbf{Simulation Testing Manager} deploys compositions by mapping them into a simulation representative of the target system. Control of the simulation can run the composition through testing phases with the simulated system under any type of definable conditions.

\noindent 4. \textbf{Deployment Manager} deploys compositions by mapping them 1:$n$ into running instances, each with its own configuration and connected to a target system. For example, we can assign a location (e.g., DNS name or IP number) and an authentication token as minimal configurations. The deployment manager continually monitors, terminates, and provides other features expected from cloud-deployed services with minimal downtime.

Composite AI systems are designed to be system-agnostic---they function the same way regardless of whether they're connected to the actual target system or a simulation of it. For testing purposes, users can connect their compositions to existing simulators (if available) or use configurable generic simulators that mimic the target system's behavior.



\section{Target System Application}
We describe an approach for building a general hierarchical autonomous control architecture using a composition. The approach combines well-known techniques from satisfiability theory, automated planning, automated diagnosis, constraint programming, reinforcement learning, and behavior trees.


A user provides a \textbf{target system} that accepts actions and provides a perception of it and its world’s state. The user works with the backend system through a \textbf{user interface} to configure one or more domain models and a composite system of AI-based components organized by a graph (e.g., behavior tree) running through tests and deployment to solve problems through action-space control.

As a running example, we will describe the components of a specific case study of this system that has been set up to solve the Towers of Hanoi game as the target system. The Towers of Hanoi is a puzzle game that involves moving discs from one place to another. Imagine you have $n$ rods, where $n$ is typically three, and a stack of different-sized discs. The goal is to move all the discs from one rod to another, following a few rules. First, you can only move one disc at a time. Second, you can never put a bigger disc on the top of a smaller one. And third, you can use the other rods to help you move the discs around, but you can only place a disc on top of another disc or on an empty rod.

The \textbf{deployed composition} contains two complementary models: a symbolically trained Towers of Hanoi Model that captures domain knowledge in PDDL format, and a Reinforcement Learning (RL) Move Evaluator trained on historical gameplay data to predict move quality scores. The models were learned by taking raw logs from a system, an expert's domain knowledge, existing domain knowledge (e.g., existing PDDL descriptions), and other sources that provide domain knowledge through capture collecting into a knowledge base. Knowledge capture primarily involves preparing a set of predicates that describe the domain under interest following any method of knowledge capture from a consultant or domain expert. A predicate is a function of a set of parameters that returns a Boolean as an answer. The knowledge base may also contain a description of predicates and their meaning extracted from the target system's raw logs. Predicates apply to a specific type of object, or to all objects. Predicates are either true or false at any point in a plan and when not declared, they are assumed to be false. In our running example, the knowledge base may contain information such as the definition of useful predicates: $discONdisc(disc1, disc2)$ means $disc1$ is on $disc2$, $smaller(disc1, disc2)$ means $disc1$ is smaller than $disc2$, $discClear(disc1)$ means $disc1$ is on the top and nothing lies on it. It may also define the goal to be achieved, such as all discs lie on the last rod, composed of predicates that are all true (e.g., $discONrod(disc1, rod3)$, $discONrod(disc2, rod3)$, and $discONrod(disc3, rod3)$). It may also define the initial state of the problem using predicates that describe the input scenario and all combinations of disc and their sizes and location.

Information from the \textbf{target system} flows into and out of the \textbf{deployed composition} through the data connectors. This takes perception data from the external world state of the system or other systems that may accurately report data from the environment of the target system. This could be any external information relevant to the domain of interest, such as weather, traffic conditions, general knowledge from the outer world, etc. The target system may provide telemetry or operational logs, such as a raw stream/batch of data collected from system components. For our running example, keys pressed in the game by a human player. The composition output data connector transmits action information to the system—transformed planned actions that lead to actual system change. In the planned action for our running example, an action can be $MoveDiscFromRodToRod$, which translated to the target system can mean ''press arrow up, press arrow left, press arrow left, press arrow left, press arrow down'' as a sequence of virtual keyboard key presses.

A data connector takes in data in some form, digests the data from the source (e.g., document, Google Drive, SFTP, stream, websocket, Parquet, AWS), and passes it in a defined structured data format to be processed by other entities. In our running example, this could involve using Keboola or some data platform to interchange data from the user's game into a format defined by the predicate store. The configuration may include format specifications, connectors, logins, secrets, etc. A data connector takes in predicate definition raw data, and transforms it given the predicate definition creating predicates from the input log stream. In our running example, $Rod1 = [disc3, disc2]$ is converted to $discONrod(disc3, rod1)$, $discONrod(disc2, rod1)$, $discONdisc(disc2, disc3)$, $smaller(disc2, disc3)$, $discClear(disc2)$. 

Within the \textbf{deployed composition}, the PDDL model provides guaranteed valid moves and optimal planning capabilities, while the RL model contributes learned heuristics about move efficiency and common solution patterns. The input data connector passes state information into a partially permeable control node that implements sophisticated orchestration logic. This control node transforms the raw game state into two formats (PDDL predicates for the planner and feature vectors for the RL model), orchestrates parallel execution of both models, combines outputs by using the planner's valid moves as candidates and the RL model's scores to rank them, and selects the highest-scored valid move as output, falling back to the planner's first suggestion if RL scoring fails. 


A data connector takes information from the composition output and converts it to an action description with parameters, which describes actions to be executed in the target system to achieve the desired goal. In our running example, something like $move(d1, d2, d3)$ means to move $disc1$ from $disc2$ to $disc3$. The output data connector interfaces with the target platform to execute all action commands on that platform, thus affecting the state through action.

The \textbf{testing and deployment user-interface} is the front-end, interactive tool that works with the user. The interface again consists of four main parts as illustrated in Figure~\ref{fig:end-to-end}:

\noindent 1. \emph{Model development} using a text-based language augmented by a visual programming language. In one case study, the modeling language is PDDL text and Blockly provides a visual builder. The user iteratively refines the model through testing and deployment in a digital simulation of the target system and/or the target system. Model debugging tools also provide insight and feedback for iterative refinement.

Models may be completely synthesized from raw observational data using a method such as the Automated Domain-driven Model Synthesis \cite{balyo2024planning}. The interface tools can then be used to manipulate, edit, add, and remove actions within the synthesized model. The goal is a model containing information from domain synthesis on which action to use or not to use for control with the target system. User augmentation may refine the model and improve understandability by renaming predicates, goals, and actions. 
    
\noindent 2. \emph{Composition development} using a text-based language augmented by a visual programming language. In one case study, the composition language is text describing a behavior tree and Blockly provides a visual builder. A behavior tree visualizer also provides additional insights. The user iteratively refines the composition and model(s) through testing and deployment in a digital simulation of the target system and/or the target system. Composition debugging tools also provide insight and feedback for iterative refinement.

The goal is given one or more models, data connectors, and control nodes, structure the composition workflow in a way that all components cooperate and achieve the desired system performance.


In our example, the composition obtains the puzzle state, then orchestrates both the PDDL planner and RL evaluator through a control node to generate optimally-ranked move sequences, which are executed on the target game.
    
\noindent 3. \emph{Simulation testing} provides pre-deployment feedback on the composition and underlying models in action. 
    
\noindent 4. \emph{Deployment} provides the ability to push control by the composition to the target system. This includes live performance monitoring of the system in action and supports defining and observing performance through evaluation metrics.

The \textbf{Deployed Model Runtime} takes the deployment setup, description of container images, image registry, relevant services, external endpoints, API's, data locations, and the input stream of target system predicates to deploy the model on a cloud-based platform. It can also run on any compute platform. It will run services and let them communicate according to description given by composition and ingest input data (pre-processed predicates) and output data (actions that lead to goal). The system output is a stream of actions to be executed on the target system. It is up to the target system to make the actions happen in the real world given action input. In our running example, this system could be a low-level Azure cloud deployment configurations of individual composition components (planner, data store, serving APIs), networking and interoperability settings, VPNs, container registry settings, endpoints definition, etc. needed to startup and execute a composed AI system to play the Towers of Hanoi. 

\section{Conclusion}
In conclusion, this paper presents a comprehensive framework for Composite AI systems that effectively integrates multiple AI techniques to solve complex real-world problems. The graph-based architecture, featuring specialized control nodes, enables the orchestration of diverse models including planning, machine learning, constraint programming, and large language models within a unified operational schema. By supporting hierarchical, sequential, and concurrent model composition patterns, the system overcomes the limitations of single-model approaches while providing flexible deployment options through containerization. The methodology encompasses data connectors, computational services, and control mechanisms that together form a powerful platform for designing, testing, and deploying intelligent automation services. As demonstrated through case studies like the Towers of Hanoi example, this approach enables the creation of robust composite systems that can observe, reason about, and act upon complex environments, representing a significant advancement in applied AI for industrial applications.

\appendix

\bibliographystyle{named}
\bibliography{citations}

\end{document}